\documentclass{article}

\usepackage[preprint]{neurips_2024}

\usepackage[utf8]{inputenc} % allow utf-8 input
\usepackage[T1]{fontenc}    % use 8-bit T1 fonts
\usepackage{hyperref}       % hyperlinks
\usepackage{url}            % simple URL typesetting
\usepackage{booktabs}       % professional-quality tables
\usepackage{amsfonts}       % blackboard math symbols
\usepackage{nicefrac}       % compact symbols for 1/2, etc.
\usepackage{microtype}      % microtypography
\usepackage{xcolor}         % colors
\usepackage{amsmath}
\usepackage{cite}
\usepackage{graphicx}

\title{Multiplicative-Additive Constrained Models:\\Toward Joint Visualization of Interactive and Independent Effects}

\author{%
  Fumin Wang\thanks{Use footnote for providing further information
    about author (webpage, alternative address)---\emph{not} for acknowledging
    funding agencies.} \\
  Independent Researcher\\
}

\begin{document}

\maketitle

\begin{abstract}
Interpretability is one of the considerations when applying machine learning to high-stakes fields such as healthcare that involve matters of life safety.  
Generalized Additive Models (GAMs) enhance interpretability by visualizing shape functions. 
Nevertheless, to preserve interpretability, GAMs omit higher-order interaction effects (beyond pairwise interactions), which imposes significant constraints on their predictive performance. 
We observe that Curve Ergodic Set Regression (CESR), a multiplicative model, naturally enables the visualization of its shape functions and simultaneously incorporates both interactions among all features and individual feature effects. 
Nevertheless, CESR fails to demonstrate superior performance compared to GAMs. 
We introduce Multiplicative-Additive Constrained Models (MACMs), which augment CESR with an additive part to disentangle the intertwined coefficients of its interactive and independent terms, thus effectively broadening the hypothesis space. 
The model is composed of a multiplicative part and an additive part, whose shape functions can both be naturally visualized, thereby assisting users in interpreting how features participate in the decision-making process. 
Consequently, MACMs constitute an improvement over both CESR and GAMs. 
The experimental results indicate that neural network–based MACMs significantly outperform both CESR and the current state-of-the-art GAMs in terms of predictive performance.
\end{abstract}

\section{Introdution}

Neural networks (NNs) have become a cornerstone of artificial intelligence (AI), with wide-ranging applications in areas such as computer vision \citep{Li_2024_CVPR} and natural language processing \citep{guo2024deepseek}. 
Despite their strong predictive capabilities, NNs are inherently complex high-dimensional function approximators, making their decision-making processes opaque and difficult to interpret. 
As a result, these models are often considered "black boxes." 
In high-stakes domains like healthcare, this lack of transparency poses serious challenges: when errors occur, it can be difficult to identify their sources or apply targeted corrections, undermining user trust and limiting reliability. 
Moreover, in fields such as biology, interpretability can be even more valuable than predictive accuracy, providing novel insights and guiding scientific discovery \citep{novakovsky2023obtaining}.

To address the opacity of black-box models, the field of Explainable AI (XAI) has received increasing attention, resulting in a rapid expansion of related literature and techniques alongside the growth of deep learning \citep{arrieta2020explainable}. However, most XAI methods provide post-hoc explanations, attempting to approximate an inherently uninterpretable model using surrogate models, gradients, feature importance measures, or other statistical tools \citep{dwivedi2023explainable}. For instance, saliency maps in computer vision highlight regions or pixels that strongly influence the model’s output \citep{chen2019looks}. While they offer some insight, saliency maps often fail to reveal how inputs are actually used in decision-making and can be unreliable, as different methods produce inconsistent results. In general, post-hoc explanations approximate the model rather than uncovering its true logic, and this approximation may compromise accuracy, potentially introducing misleading interpretations in high-stakes scenarios \citep{lakkaraju2020fool}.

In contrast, Interpretable Machine Learning (IML) focuses on designing models whose decision-making processes are inherently understandable and trustworthy \citep{rudin2022interpretable}. Examples include generalized linear models \citep{nelder1972generalized}, score-based methods \citep{ustun2016supersparse}, and decision trees \citep{quinlan1986induction}. These models are typically structured to satisfy desirable properties such as monotonicity, additivity, sparsity, causality, or other domain-specific constraints \citep{carvalho2019machine}. Generalized Additive Models (GAMs) \citep{hastie2017generalized}, as an extension of generalized linear models, capture nonlinear relationships between predictors and outcomes through flexible shape functions $f_i(*)$, typically univariate and visualizable. GAMs can be expressed as:
\begin{equation}\label{1}
g(E[y]) = \beta + f_{1}(x_{1}) + f_{2}(x_{2}) + \dots + f_{k}(x_{k}),
\end{equation}
where $g(\cdot)$ is the link function, $X=(x_{1},\dots,x_{k})$ represents the independent variables, $y$ is the dependent variable, and each $f_i(\cdot)$ is a shape function that facilitates interpretability by explicitly showing each feature’s contribution. Although GAMs are a long-established technique, their inherent interpretability has recently rekindled interest, motivating the development of approaches such as GAMI-Net \citep{yang2021gami}, Sparse Interaction Additive Networks \citep{enouen2022sparse}, and other neural-network-based extensions. These methods primarily focus on incorporating pairwise interactions or combining GAMs with neural networks to balance interpretability and predictive accuracy, setting the stage for further advances in interpretable modeling.
However, traditional additive models are typically limited to modeling independent effects and often fail to capture complex feature interactions. Even GA$^{2}$Ms, which incorporate pairwise interactions, are unable to represent interactions beyond two variables. This restricts their expressiveness, particularly in domains where such interactions are critical to understanding the underlying processes. 

Building on recent advances, particularly the proposed Curve Ergodic Set Regression (CESR) \citep{wang2022ergodic}, we investigate an alternative formulation that seeks to address the limitations of additive models. CESR is expressed as:
\begin{equation}\label{CESR}
y_{C} = C \cdot \prod_{i=1}^{k}\left(1 + w_{i1} x_{i}^{1} + \cdots + w_{i n_{i}} x_{i}^{n_{i}}\right)= C \cdot U_{1}(x_{1}) \cdot U_{2}(x_{2}) \cdots U_{k}(x_{k}),
\end{equation}
where each $U_i(\cdot)$ is a univariate shape function. By construction, CESR naturally incorporates both independent and higher-order interaction effects, which in principle should enable stronger predictive performance than GAMs. However, in practice, its effectiveness is limited because the coefficients of the independent terms and the interaction terms are entangled rather than separated, leading to instability during training and reduced performance.

In this paper, we propose Multiplicative-Additive Constrained Models (MACMs), which jointly determines outcomes through both multiplicative and additive parts. This model has the following characteristics:

$\bullet$MACM is a constrained model designed to remain interpretable while achieving performance as close as possible to its full-complexity counterpart.

$\bullet$MACM is a novel interpretable machine learning approach that considers both the independent effects of each feature on the outcome and the interactions among all features-not just pairwise interactions.

$\bullet$Compared to the CESR, the introduction of additive part decouples the coefficients of independent and interaction terms, while also expanding the model’s hypothesis space—thereby enhancing both accuracy and expressiveness.

$\bullet$Both the multiplicative and additive parts of the model can be visualized as graphs, where their product or sum comprehensively represents the decision-making process. Unlike traditional additive models, which solely capture independent effects of features, the proposed model introduces a novel perspective. It enables users to examine, from a global viewpoint, how the interactions among all features collectively influence the model’s output. 

$\bullet$The shape functions in this model are not restricted to polynomials; they can be any functional mappings. In this paper, we further introduce neural networks based MACMs by employing full connected layer as shape functions, enhancing the model's expressive power and accuracy.

\section{Related Works}
As Ergodic Set Regression (ESR) and Curve Ergodic Set Regression (CESR) constitute the cornerstone and motivation of this work, this section presents a concise review and description of these relatively niche models.
\subsection{Ergodic Set Regression}
Ergodic Set Regression (ESR) \citep{wang2022ergodic} is an enhanced multivariate polynomial regression (MPR) model \citep{article}, constructed on a specialized set of polynomial power terms $[E_{N}(X)]$, where $X$ denotes the input vector and $N$ is the maximum polynomial degree. All elements in this set are generated from the following function:
\begin{equation}\label{Efunc}
\begin{aligned}\vspace{1.5ex}\displaystyle
\left(1+x_{1}^{1}+\cdots+x_{1}^{n_{1}}\right) \cdot
\left(1+x_{2}^{1}+\cdots+x_{2}^{n_{2}}\right) \cdot \cdots \cdot
\left(1+x_{k}^{1}+\cdots+x_{k}^{n_{k}}\right).
\end{aligned}
\end{equation}

The general form of ESR can then be written as:
\begin{equation}\label{ESR}
\begin{aligned}\vspace{1.5ex}\displaystyle
ESR: [W_{lin}] \times \left[E_{N}(X)\right]^{T},
\end{aligned}
\end{equation}
where $[W_{lin}]$ denotes the linearized coefficients. Owing to the specific construction of $[E_{N}(X)]$, ESR leverages a more comprehensive set of polynomial terms compared with standard MPR under the same polynomial degree, thereby achieving higher accuracy. For example, given $X=(x_{1}, x_{2})$, when the maximum polynomial degree is $(1,1)$ for ESR and $1$ for MPR, the term expansion in ESR is $[1, x_{1}, x_{2}, x_{1}x_{2}]$, while that in MPR is limited to $[1, x_{1}, x_{2}]$.

Despite both ESR and MPR residing in high-dimensional function spaces, ESR exhibits an inherent structural property that allows it to be readily transformed into an interpretable model.

\subsection{Curve Ergodic Set Regression}
Curve Ergodic Set Regression (CESR) is a constrained variant of ESR. Similar to ESR, CESR employs $[E_{N}(X)]$ as the basis for model construction; however, it introduces nonlinearity through its parameterization. By examining \eqref{Efunc}, one can naturally associate it with the following nonlinear coefficients derived from the following form:
\begin{equation}\label{nonli_coe}
\begin{aligned}\vspace{1.5ex}\displaystyle
\left(1+w_{1,1}+\cdots+w_{1,n_{1}}\right) \cdot
\left(1+w_{2,1}+\cdots+w_{2,n_{2}}\right) \cdot \cdots \cdot
\left(1+w_{k,1}+\cdots+w_{k,n_{k}}\right).
\end{aligned}
\end{equation}

When the nonlinear coefficient matrix is applied, the model will be interpretable, and CESR can be expressed as:
\begin{equation}\label{CESR}
\begin{aligned}\vspace{1.5ex}\displaystyle
CESR&:[W_{nonli}] \times\left[E_{N}(X)\right]^{T} \\
&= C \cdot \prod_{i=1}^{k}\left(1+w_{i1}x_{i}^{1}+\cdots+w_{i n_{i}} x_{i}^{n_{i}}\right) \\
&= C \cdot U_{1}(x_{1}) \cdot U_{2}(x_{2}) \cdot \cdots \cdot U_{k}(x_{k}),
\end{aligned}
\end{equation}
where $U_{i}(x_{i})$ denotes the shape function of feature $x_{i}$. Each $U_{i}(\cdot)$ can be visualized as a curve, enabling intuitive interpretation of both individual contributions and the overall decision-making process. Notably, $U_{i}(\cdot)$ always passes through $(0,1)$. This property arises because the bias of $U_{i}(\cdot)$ is extracted outside the function, so that the constant $C$ becomes the product of all biases. Consequently, $C$ serves to normalize the shape functions, avoiding misleading model behavior and eliminating the issue of infinitely many identical local minima that could otherwise obscure interpretation.

Expanding \eqref{CESR} while preserving all terms but reordering them yields:
\begin{equation}\label{excesr}
\begin{aligned}\vspace{1ex}\displaystyle
C \cdot [1 &+ (w_{11} x_{1}^{1}+ \cdots+w_{1 n_{1}} x_{1}^{n_{1}})+\cdots+(w_{k1} x_{k}^{1}+\cdots+w_{k n_{k}} x_{k}^{n_{k}}) \\
&+ w_{11}w_{21} x_{1}^{1}x_{2}^{1}+\cdots+ \prod_{i=1}^{k}(w_{i n_{i}} x_{i}^{n_{i}})]
\end{aligned}
\end{equation}
In essence, as shown in \ref{excesr}, CESR can be regarded as a constrained subset of ESR: it remains a complex polynomial model but achieves interpretability by imposing nonlinear structure on its parameters. 
CESR provides users with a new perspective: it illustrates the multiplicative effect of all factors on the prediction while simultaneously retaining the power terms contained in polynomial based GAMs
By contrast, in GAMs, achieving interpretability typically requires discarding all interaction terms.

\begin{table}[h]
\centering
\begin{minipage}{0.6\textwidth}
\renewcommand\arraystretch{1.5} 
As shown in \eqref{excesr}, CESR incorporates not only multiplicative components (interaction terms) but also additive components (independent terms) that constitute GAMs. 
Theoretically, this enables CESR to represent more complex functions than polynomial based GAMs, thereby offering higher potential accuracy. 
However, due to the nonlinear parameterization, the coefficients of additive and multiplicative components are inherently coupled. 
For instance, $w_{1}$ in the additive component serves as the coefficient of $x_{1}$ but simultaneously contributes to the coefficients of interaction terms involving $x_{1}$, such as $x_{1}x_{2}, x_{1}x_{2}x_{3}, \ldots, \prod_{i=1}^{k}(x_{i}^{n_{i}})$. 
Such coupled coefficients prevent the hypothesis space of CESR from encompassing that of GAMs.
In practice, as demonstrated in Table \ref{fig:pidd}, this coupling often prevents CESR from outperforming polynomial based GAMs.
\end{minipage}
\hfill 
\begin{minipage}{0.35\textwidth}
\centering
\caption{AUC on Pima Indian Diabetes Dataset\citep{pidd}. Under the condition that the maximum polynomial degree is set to 7, the results are obtained through 5-fold cross-validation.}
\label{fig:pidd}
\renewcommand\arraystretch{1.5} 
\begin{tabular}{cccc}
\hline
model   &      \ \ \ \ \ \ AUC   \\    \hline
CESR                      & \ \ $0.8493\pm0.154$       \\
GAMs(Poly)           & \ \ $0.8533\pm0.117$       \\
\hline
\end{tabular}
\end{minipage}
\end{table}

\section{Multiplicative-Additive Constrained Models}
\label{gen_inst}

\subsection{Decoupling of Coefficients and the Expanded Hypothesis Space }
To decouple the coefficients of independent and interaction terms in CESR, we introduce an additive polynomial into the model. The resulting formulation consists of a multiplicative part and an additive part:
\begin{equation}\label{origin}
\begin{aligned}\vspace{1.5ex}\displaystyle
\prod_{i  =  1}^{k}\left(w_{i0}^m+w_{i 1}^m x_{i}^{1}+\cdots+w_{i n_{i}}^{m} x_{i}^{n_{i}}\right)+\sum_{i = 1}^{k}(w_{i 0}^{a}+w_{i 1}^{a} x_{i}^{1}+\cdots+w_{i n_{i}}^a x_{i}^{n_{i}}),
\end{aligned}
\end{equation}
where $w_{ij}^m$ and $w_{ij}^a$ denote the coefficients of the multiplicative and additive parts, respectively. 
The polynomial order of each feature is kept consistent across the two parts, but their coefficients are parameterized independently. 
Expanding this model yields:
\begin{equation}\label{exp_ori}
\begin{aligned}\vspace{1.5ex}\displaystyle
&(\prod_{i=1}^{k}(w_{i 0}^{m})+\sum_{i=1}^{k}(w_{i 0}^{a}))\\
&+(w_{11}^{m}+w_{11}^{a}) x_{1}^{1}+ \cdots+(w_{1 n_{1}}^{m}+w_{1 n_{1}}^{a}) x_{1}^{n_{1}}+\cdots +(w_{k n_{1}}^{m}+w_{k n_{k}}^{a}) x_{k}^{n_{k}}\\
&+ w_{11}^{m}w_{21}^{m} x_{1}^{1} x_{2}^{1}+\cdots+ \prod_{i=1}^{k}(w_{i n_{i}}^{m} x_{i}^{n_{i}})
\end{aligned}
\end{equation}
Here, the first row represents the constant term, the second row corresponds to the independent terms, and the third row captures the interaction terms. 
Compared with \eqref{excesr}, the additive part introduces free coefficients that adjust the independent terms, thereby decoupling them from the interaction terms. 
This decoupling broadens the hypothesis space and enhances the representational capacity of the model.
To illustrate, consider the target function:
$$
y=1+x_{1}+(1+k)x_{2}+x_{1}x_{2},k\neq0.
$$
For CESR (or the multiplicative part alone) $\prod_{i = 1}^{2}(w_{i 0}^{a}+w_{i 1}^{a} x_{i})$ to approximate $y$ without error, it must satisfy:
$$
w_{10}^{m}w_{20}^{m}=1,w_{11}^{m}w_{20}^{m}=1,w_{11}^{m}w_{21}^{m}=1,
$$
which further implies $w_{10}^{m}w_{21}^{m}=1$. Thus, the expansion of the multiplicative part becomes $1+x_{1}+x_{2}+x_{1}x_{2}$, leading to an error of $|kx_{2}|$. Since the multiplicative part cannot capture this discrepancy, CESR fails to exactly represent $y$, as well as the additive part which alone cannot model the interaction term $x_{1}x_{2}$.
By incorporating the additive part $\sum_{i = 1}^{2}(w_{i 0}^{a}+w_{i 1}^{a} x_{i})$ and setting $w_{10}^{a}=w_{11}^{a}=w_{21}^{a}=0, w_{20}^a=k$, the full model becomes $1+x_{1}+x_{2}+x_{1}x_{2}+kx_{2}$, which exactly matches $y$. Hence, the additive part serves to decouple and correct coefficients, enabling the proposed model to represent functions that CESR cannot.

From another perspective, the introduction of the additive part enlarges the hypothesis space of the multiplicative part. Let
$$
\mathcal{H}_{m}=\{ \prod_{i  =  1}^{k}\left(w_{i0}^m+w_{i 1}^m x_{i}^{1}+\cdots+w_{i n_{i}}^{m} x_{i}^{n_{i}}\right)|x_{i} \in \mathbb{R}^{k},k\geq2 \}
$$ 
and 
$$\mathcal{H}_{a}=\{ \sum_{i  =  1}^{k}\left(w_{i0}^a+w_{i 1}^a x_{i}^{1}+\cdots+w_{i n_{i}}^{a} x_{i}^{n_{i}}\right)|x_{i} \in \mathbb{R}^{k},k\geq2 \}$$  
represent the hypothesis spaces of the multiplicative and additive part, respectively. 
When the number of features $k \geq 2$, $\mathcal{H}_{m}$ and $\mathcal{H}_{a}$ have little intersection. Consider the following equation:
$$
\prod_{i  =  1}^{k}\left(w_{i0}^m+w_{i 1}^m x_{i}^{1}+\cdots+w_{i n_{i}}^{m} x_{i}^{n_{i}}\right) =  \sum_{i  =  1}^{k}\left(w_{i0}^a+w_{i 1}^a x_{i}^{1}+\cdots+w_{i n_{i}}^{a} x_{i}^{n_{i}}\right).
$$
To ensure formal equivalence, for any non-zero coefficient $w_{q v}^{a}$ of $x_{q}^{v}(v\neq 0)$, one necessary condition is
$$
w_{q v}^{m}=w_{q v}^{a}, w_{q v}^{m}*w_{i j}^m=0,i \neq q,i \neq 0,j=0,\cdots,n_{c}.
$$
This implies $w_{ij}^{m}=0$ for all $i\neq q$, which means the left-hand side reduces to a univariate function, i.e., $k=1$, contradicting the assumption $k \geq 2$. Therefore,
$$
dim(\mathcal{H}_{m}+\mathcal{H}_{a})>max\{dim(\mathcal{H}_{m}),dim(\mathcal{H}_{a})\}.
$$
Hence, $\mathcal{H}_{m}$ and $\mathcal{H}_{a}$ have little intersection when $k \geq 2$, and the essence of combining the multiplicative and additive part is to expand the hypothesis space of CESR or polnomial based GAMs, thereby granting their combination greater expressive power in theory.

\subsection{The Form of Multiplicative-Additive Constrained Models}
Unlike neural networks, which possess theoretically universal approximation capabilities \citep{256500}, polynomials cannot approximate arbitrary functions with arbitrary precision \citep{9def9caf-331b-3666-a9d2-8034c4dde74a}. 
This limitation inherently constrains the performance of Equation~\eqref{origin}. 
To overcome this, we generalize Equation~\eqref{origin} by removing the restriction to polynomial bases. 
The resulting framework is referred to as Multiplicative-Additive Constrained Models (MACMs), defined as:
\begin{equation}\label{mam}
\begin{aligned}\vspace{1.5ex}\displaystyle
MACMs:\prod_{i=1}^{k} f_{mi}(x_i) + \sum_{i=1}^{k} f_{ai}(x_i),
\end{aligned}
\end{equation}
where $f_{mi}:\mathbb{R}\rightarrow\mathbb{R}$ and $f_{ai}:\mathbb{R}\rightarrow\mathbb{R}$ denote the feature shape functions of the multiplicative and additive part, respectively. Importantly, these shape functions are not limited to polynomials and may be instantiated by arbitrary functional mappings. 
Since $f_{mi}$ and $f_{ai}$ are both univariate, they can be directly visualized as curves, providing an intuitive interpretation of how each feature contributes multiplicatively and additively to the prediction. 
Furthermore, the combination of these curves (via product and sum) faithfully reflects the model’s decision-making process, thereby preserving interpretability while enhancing flexibility.

\subsubsection{Interpretability of MACMs}
The interpretability of MACMs arises from two key aspects: the visualizable decision process and the dynamic contribution of each feature to the output. However, because MACMs involve multiplicative operations, large-magnitude feature values may cause gradient explosion during training. To mitigate this, input features are normalized prior to training using min–max scaling to the range $[-1,1]$. Training is then performed on the normalized data, and the feature shape functions are mapped back to the original scale through a linear transformation after training. 

Unlike CESR in Equation~\eqref{CESR}, the multiplicative part of MACMs does not extract constants from each $f_{mi}$. While the two forms yield identical outputs, leaving constants inside $f_{mi}$ allows the functions to belong to $\mathcal{H} = \{ g : X \to \mathbb{R} \mid g(0) = 0 \}$. For example, $1+w_{i1}x_i+\cdots+w_{in_i}x_i^{n_i}$ cannot represent $x+x^2$, whereas $w_{i0}^m+w_{i1}^mx_i+\cdots+w_{in_i}^mx_i^{n_i}$ can. However, this choice may also yield misleading interpretations \citep{wang2022ergodic}. Consider two models: 
\[
y_{1}=\bigl(10^5+10^5x_{1}+10^5x_{1}^2\bigr)(1+x_{2}+x_{2}^2),\quad 
y_{2}=\bigl(1+x_{1}+x_{1}^2\bigr)(10^5+10^5x_{2}+10^5x_{2}^2).
\]
Although $y_1$ and $y_2$ produce identical outputs, the curves suggest that $x_1$ dominates in $y_1$ and $x_2$ dominates in $y_2$, leading to contradictory interpretations. To resolve this, we extract constant terms for normalization. The model is transformed into the following form for visualization:
\begin{equation}\label{mam-trans}
\begin{aligned}\vspace{1.5ex}\displaystyle
MACMs &: \prod_{i=1}^{k}f_{mi}(0)\prod_{i=1}^{k}\frac{f_{mi}(x_i)}{f_{mi}(0)} 
+ \sum_{i=1}^{k}f_{ai}(0) + \sum_{i=1}^{k}\bigl(f_{ai}(x_i)-f_{ai}(0)\bigr) \\[0.5ex]
&= C_{m}\prod_{i=1}^{k}U_{mi}(x_i) + C_{a} + \sum_{i=1}^{k}U_{ai}(x_i),
\end{aligned}
\end{equation}
where $U_{mi}$ and $U_{ai}$ are normalized shape functions of the multiplicative and additive parts, respectively. By construction, $U_{mi}(0)=1$ and $U_{ai}$ contains no bias term. The global constant term is $C_{m}+C_{a}$. This transformation requires $f_{mi}(0)\neq0$; otherwise, constants for that feature are left unextracted. 

This normalization primarily facilitates visualization. Given these normalized functions, one can manually derive the model’s output. To describe the nonlinear relationship between a feature $x_i$ and the output, we isolate $x_i$ as the independent variable:
\[
MACMs: \alpha U_{mi}(x_i) + U_{ai}(x_i) + \beta,
\]
where $\beta=\sum_{j\neq i}U_{aj}(x_j)$ is the additive bias and $\alpha=C_{m}\prod_{j\neq i}U_{mj}(x_j)$ is a dynamic scaling factor determined by the multiplicative part. Hence, the contribution of $x_i$ is dynamic: when $\alpha$ is small, the additive component $U_{ai}(x_i)$ dominates, whereas for large $\alpha$, the multiplicative part governs the nonlinear relationship. To visualize this effect, we plot dynamic curves of the form:
\begin{equation}\label{mam-dynamic}
\alpha U_{mi}(x_i) + U_{ai}(x_i), \quad i=1,\dots,k,
\end{equation}
where $\alpha$ varies continuously. Since infinitely many such curves exist, we sample $\alpha$ uniformly from $[\min(\alpha), \max(\alpha)]$ in 10 steps, yielding a tractable set of dynamic influence curves that reveal how each feature’s relationship with the output evolves as $\alpha$ changes.

\subsection{Neural Networks Based Multiplicative-Additive Constrained models}
\begin{figure}[h]
    \centering
    \includegraphics[width=0.6\textwidth]{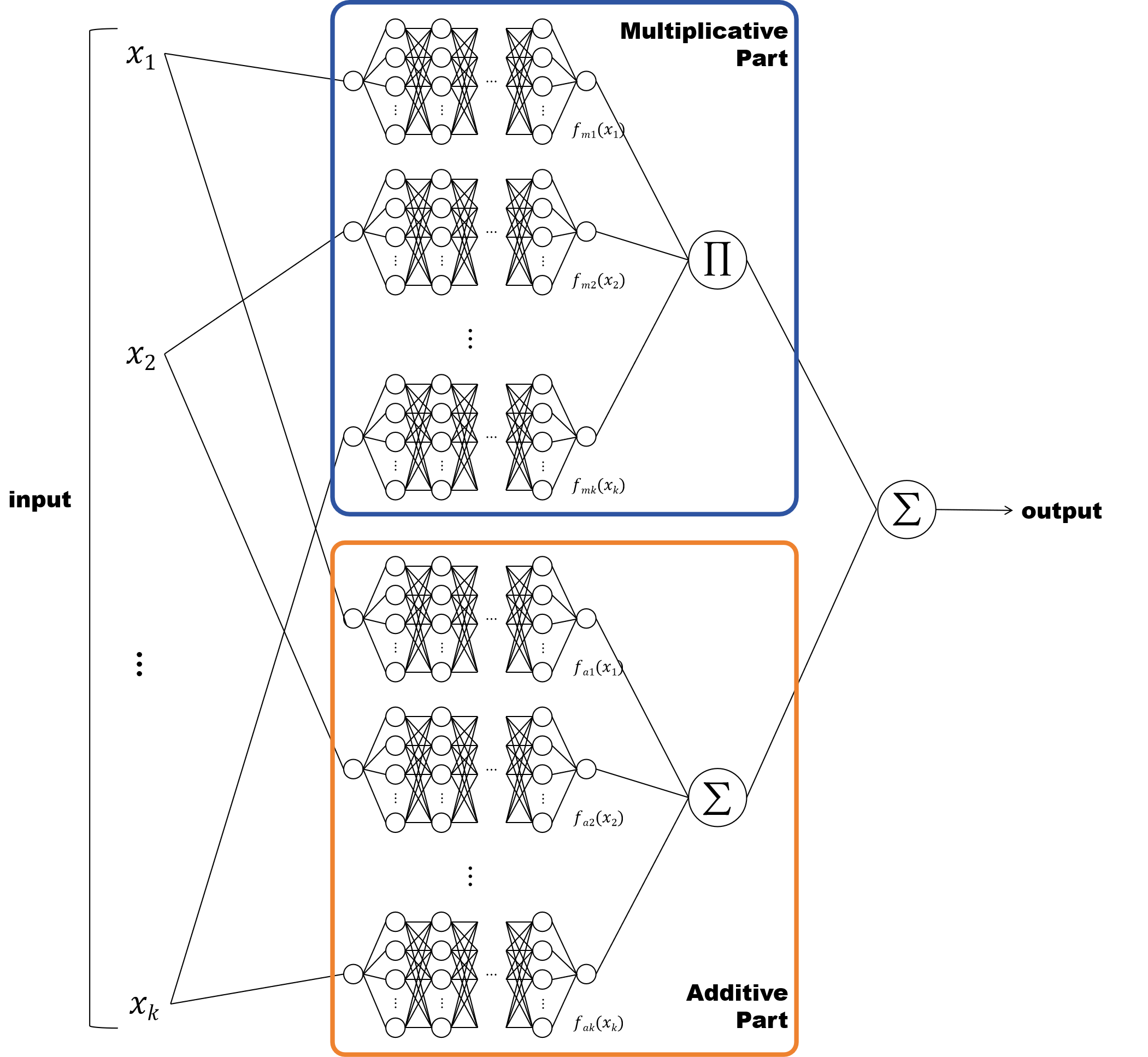}
    \caption{Architecture of MACMs(NNs), consisting of a multiplicative and an additive part. The multiplicative part is constructed as the product of multiple subnetworks, while the additive part is the sum of multiple subnetworks.}
    \label{fig:MACNNs}
\end{figure}

When the shape functions in \eqref{mam} are instantiated as fully connected neural networks, the resulting models are referred to as neural networks based Multiplicative-Additive Constrained Models (MACMs(NNs)). MACMs(NNs) serve as the primary architecture studied in this paper, and their structure is illustrated in Figure~\ref{fig:MACNNs}. 

Formally, $f_{mi}(x_i)$ and $f_{aj}(x_j)$ denote fully connected layers with arbitrary depth and width. The overall output is given by the sum of the multiplicative and additive part, which can be transformed into the form of \eqref{mam-trans} for visualization. 

During training, the parameter updates in the multiplicative part typically lag behind those in the additive part. This effect arises because subnetworks in the multiplicative part often employ bounded activation functions (e.g., ReLU), producing outputs whose absolute values do not exceed 1. Multiplying such outputs together amplifies the risk of gradient vanishing. To address this, we introduce a scaling factor $k>1$ to the multiplicative part, as shown in \eqref{macm-k}. This rescaling increases the initial gradient magnitude, thereby improving convergence speed while preserving the model’s representational capacity.  
\begin{equation}\label{macm-k}
\begin{aligned}\vspace{1.5ex}\displaystyle
MACMs(NNs): \; k \cdot \prod_{i=1}^{k} f_{mi}(x_i) + \sum_{i=1}^{k} f_{ai}(x_i).
\end{aligned}
\end{equation}
Given a training dataset $\mathcal{D}=\{(x_i, y_i)\}_{i=1}^{n}$ with input features $x \in \mathbb{R}^{k}$ and targets $y$, the model output is defined as  
\[
m^{\theta}(x) = k^{\theta} \cdot \prod_{i=1}^{k} f_{mi}^{\theta}(x_i) + \sum_{i=1}^{k} f_{ai}^{\theta}(x_i).
\]
For regression tasks, MACMs(NNs) employ the root mean squared error (RMSE) loss:
\[
l(x,y;\theta) = \sqrt{\tfrac{1}{n} \sum_{i=1}^{n} (m^{\theta}(x_i) - y_i)^2},
\]
and for binary classification, the cross-entropy loss:
\[
l(x,y;\theta) = -y \log(\sigma(m^{\theta}(x))) - (1-y) \log(1 - \sigma(m^{\theta}(x))).
\]
The overall training objective is then given by
\[
\mathcal{L}(\theta) = \mathbb{E}_{(x,y)\in \mathcal{D}} \big[l(x,y;\theta)\big].
\]

\section{Evaluation}

\subsection{Basellines}
To evaluate the performance of MACMs and demonstrate how interpretability can be achieved, we compare MACMs(NNs) and polynomial-based MACMs (MACMs(poly)) against the following baseline models:

\textbf{Neural Additive Models (NAMs).}  
A neural networks based GAMs, where each input feature is processed through an exp-centered unit (ExU) before the activation function. NAMs have shown performance comparable to Explainable Boosting Machines (EBMs) \citep{agarwal2021neural}.  

\textbf{Neural Basis Models (NBMs).}  
Another neural networks based GAMs, which employs basis decomposition of feature functions. A small number of jointly learned basis functions are shared across all features, allowing NBMs to scale efficiently to high-dimensional and sparse data. NBMs have been shown to outperform NAMs in predictive accuracy \citep{radenovic2022neural}.  

\textbf{Prototypical Neural Additive Models (ProtoNAMs).}  
An extension of NAMs that incorporates prototype learning into each feature function. Prototypes represent representative samples from the data, enabling ProtoNAMs to capture more nuanced feature effects. They have been reported to outperform all existing neural network-based GAMs \citep{xiong2024protonam}.  

\textbf{Ergodic Set Regression (ESR).}  
A generalization of polynomial regression that includes all possible polynomial interactions across input variables. ESR serves as the full-complexity counterpart relative to CESR and MACMs(poly).  

\textbf{Curve Ergodic Set Regression (CESR).}  
A constrained variant of ESR that retains all interaction terms while ensuring interpretability. CESR leverages nonlinear parameterizations, with model predictions jointly determined by the product of individual effect curves and a normalization constant.  

\textbf{Deep Neural Networks (DNNs).}  
A standard fully connected network with the same number of hidden layers as MACMs(NNs). It serves as the unconstrained full-complexity counterpart, lacking the structural interpretability imposed by MACMs. 

\subsection{Datasets}
\textbf{CA Housing (modified).}  
This dataset is a modified version of the California Housing dataset \citep{pace1997sparse}. We first removed 207 samples with missing values in the \texttt{total\_bedrooms} column. An additional categorical attribute, \texttt{ocean\_proximity}, was added and encoded as (ISLAND: 1, NEAR OCEAN: 2, NEAR BAY: 3, 1H\textless OCEAN: 4, INLAND: 5). The target variable, the median house price, was scaled by dividing all values by 1000. The dataset is publicly available at \href{https://www.kaggle.com/datasets/camnugent/california-housing-prices}{Kaggle}.  

\textbf{Stroke Prediction.}  
This dataset is used to predict whether a patient is likely to have a stroke based on demographic and health-related attributes. We removed the \texttt{id} column and encoded categorical features as follows: \texttt{gender} (Female: 0, Male: 1), \texttt{ever\_married} (No: 0, Yes: 1), \texttt{work\_type} (Children: 0, Never\_worked: 0.5, Govt\_job: 0.5, Self-employed: 0.75, Private: 1), \texttt{residence\_type} (Rural: 0, Urban: 1), and \texttt{smoking\_status} (Never\_smoked: 0, Formerly\_smoked: 0.5, Smokes: 1). All missing values were removed. The prediction target is a binary variable indicating whether a patient has a stroke. The dataset is publicly available at \href{https://www.kaggle.com/datasets/fedesoriano/stroke-prediction-dataset}{Kaggle}.  

\textbf{Water Quality Prediction.}  
This dataset contains water quality measurements collected from the Ju River by the Langfang Environmental Monitoring Station of the Beijing Institute of Environmental Planning between August 18, 2018, and March 18, 2019 \citep{yan2023combining}. The task is to predict the \texttt{Total\_Nitrogen (TN)} concentration at the next time step based on historical water quality indicators.  

For all datasets, we applied only two preprocessing operations: min-max normalization and the removal of missing values. We evaluated model performance on the CA Housing (modified) and Stroke Prediction datasets, and conducted ablation studies on the Water Quality Prediction dataset.

\subsection{Implementation Details}
\textbf{MACMs(poly).}  
For all polynomial-based MACMs, the degree of each feature shape function was set to 12, and the scaling factor $k$ was set to 20. Models were trained using the Adam optimizer with a batch size of 1024, a fixed learning rate of 0.005, and 5000 epochs. No dropout or learning rate decay was applied.  

\textbf{MACMs(NNs).}  
Each feature shape function was parameterized by a fully connected network with 10 hidden layers and 20 neurons per layer, using ReLU activations. The scaling factor $k$ was set to 10 for regression tasks. Regression models were trained with a batch size of 1024, a fixed learning rate of 0.0005 with exponential decay (factor 0.99 every 100 epochs), 0 dropout, and 10000 training epochs using the Adam optimizer. For binary classification, the scaling factor was increased to $k=1000$, a sigmoid function was applied to the outputs, the learning rate was set to 0.00005 with decay factor 0.995 every 10 epochs, and models were trained for 2000 epochs; all other hyperparameters remained the same as in the regression setting.  

For all datasets, 80\% of the samples were used for training and the remaining 20\% for testing, with data randomly shuffled during loading. Model performance was evaluated using RMSE for regression and AUC for binary classification, with reported results averaged over 5-fold cross-validation.

\subsection{Performance of MACMs and Visualization}
Table \ref{table:experiment} presents the predictive performance of MACMs. MACMs(poly) achieve results intermediate between CESR and the full-complexity ESR, whereas MACMs(NNs) attain performance between the multiplicative part alone and their full-complexity counterpart, DNNs, consistent with our expectations. We further compared MACMs(NNs) against mainstream neural network-based GAMs. Since the hypothesis space of MACMs(NNs) can be regarded as the union of the multiplicative and additive hypothesis spaces, the model benefits from this enlarged space, yielding notable performance gains over NAMs, NBMs, and ProtoNAM, even without exhaustive hyperparameter tuning. This advantage is particularly pronounced in regression tasks.  

\begin{figure}[h]
  \centering
  \includegraphics[width=0.49\textwidth]{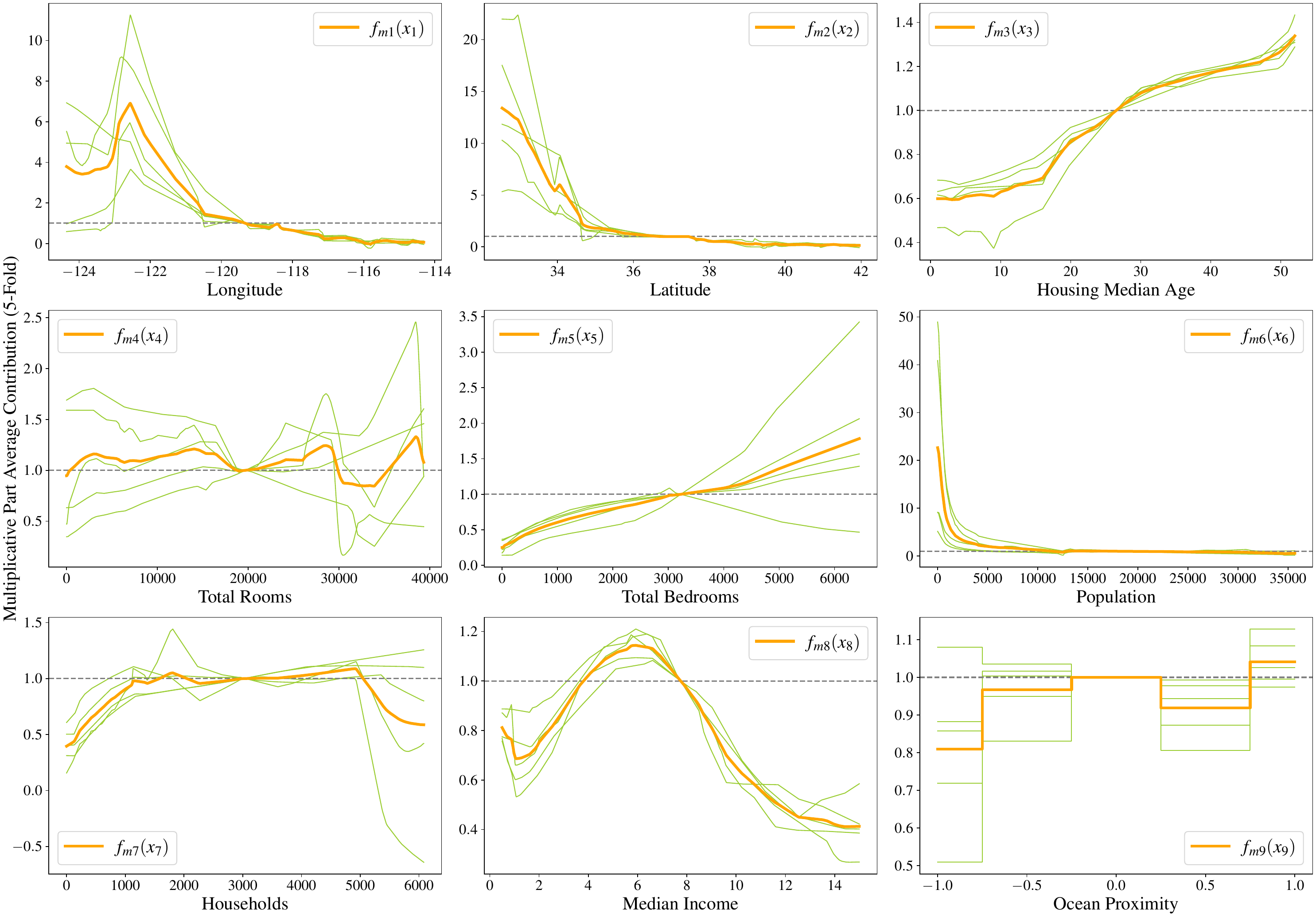}
  \includegraphics[width=0.49\textwidth]{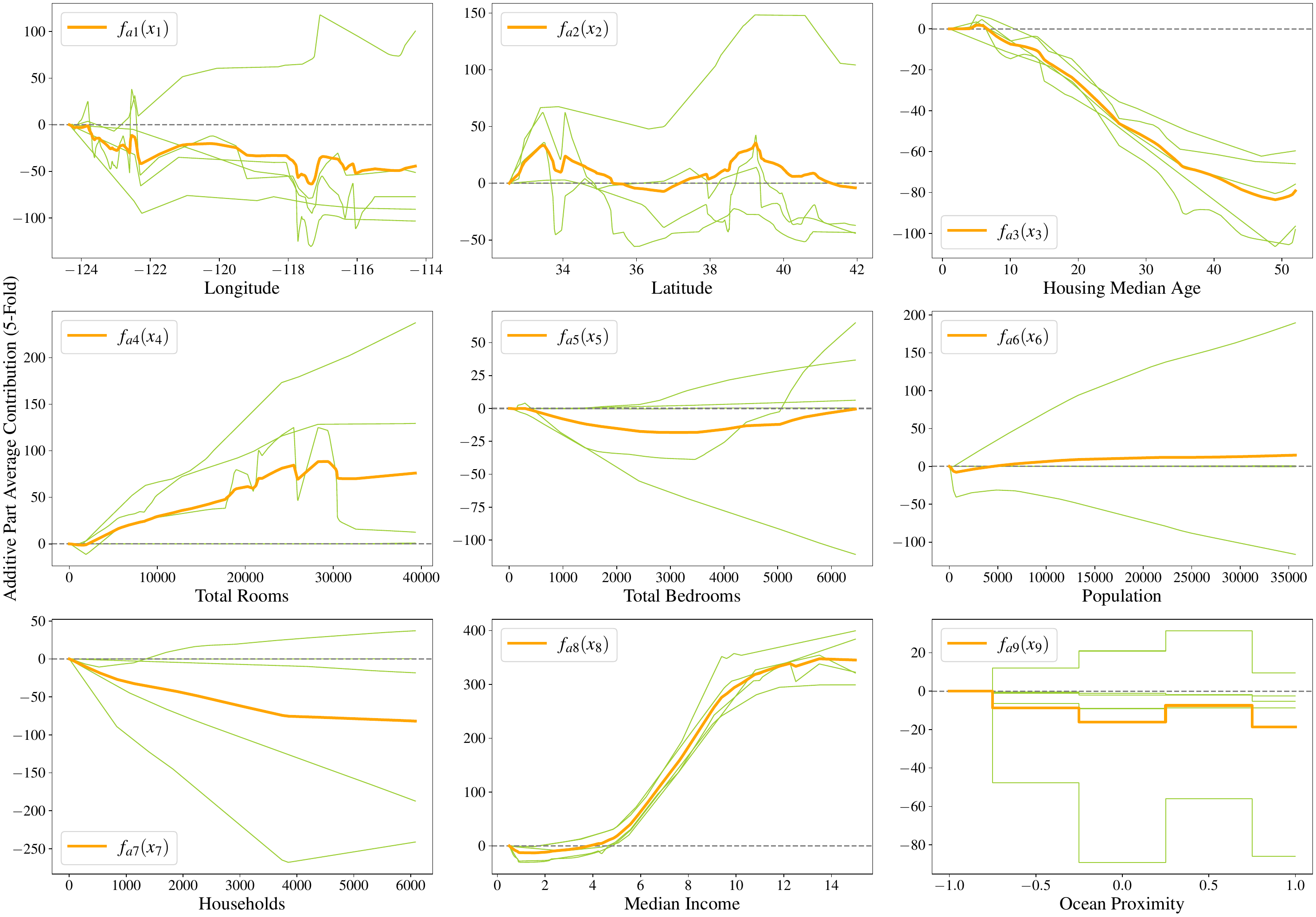}
  \caption{Multiplicative Part and Additive Part Shape Functions of MACMs(NNs) on CA Hosuing(modified).}\label{figure:mpapcam}
\end{figure}

Figure \ref{figure:mpapcam} visualizes the learned shape functions of MACMs(NNs) on the CA Housing (modified) dataset. In the plots, the green thin lines correspond to the shape functions obtained from each fold of 5-fold cross-validation, while the orange thick line represents the averaged shape function across all folds. In this experiment, with $C=51.89$ and nearly all shape function values positive, we marked the reference line $y=1$ within the multiplicative part. From a microscopic perspective, Housing Median Age and Total Bedrooms exhibit an overall positive correlation with the multiplicative part, whereas Population demonstrates a negative correlation. In contrast, within the additive part, the monotonicity often differs: Housing Median Age tends to show a negative contribution, while Median Income exhibits a positive contribution.  

\begin{figure}[h]
  \centering
  \includegraphics[width=\textwidth]{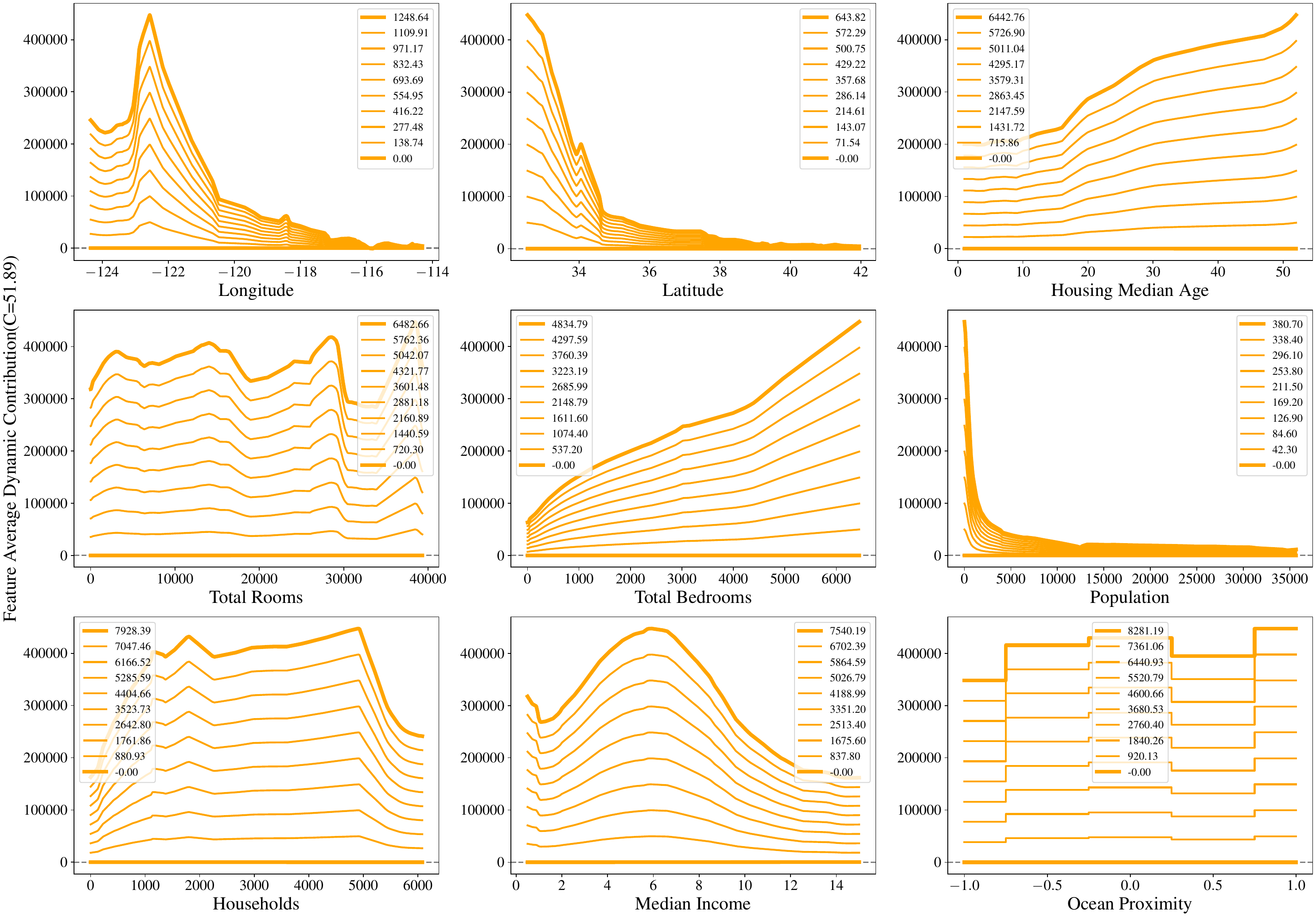}
  \caption{Multiplicative Part and Additive Part Shape Functions  of MACMs(NNs) on CA Hosuing(modified)}\label{figure:dynamiccam}
\end{figure}

Figure \ref{figure:dynamiccam} presents the dynamic shape functions of all features, following Equation \eqref{mam-dynamic}. As each feature has infinitely many dynamic curves, we display 10 uniformly sampled curves for representative illustration. The plots reveal that, although the shape functions vary dynamically, their overall forms remain structured. With increasing $\alpha$, the dynamic shape functions increasingly resemble those of the multiplicative part, illustrating how the multiplicative part progressively dominates the feature contributions.
\begin{table}[h]\renewcommand\arraystretch{1.5}\centering
\caption{Performance of MACMs and baselines. All results are reported as the mean of 5-fold cross-validation. Lower RMSE values indicate better performance, whereas higher AUC values are preferred.}\label{table:experiment}
\begin{tabular}{lccccc}
\hline
        				&CA Housing(modified)  			&CA Housing 					&Stroke  						&Water Quality 				\\ 	\hline
\bf{Model}         		&RMSE  	 					&RMSE  						&AUC  						&RMSE					\\    	\hline
NAMs      			&56.5699{\scriptsize$\pm$0.6841}	&56.4011{\scriptsize$\pm$0.6131}	&0.8190{\scriptsize$\pm$0.0227}		&0.4757{\scriptsize$\pm$0.0402}\\
CESR                		&64.7516{\scriptsize$\pm$2.8115}	&64.1922{\scriptsize$\pm$0.9096}	&0.8104{\scriptsize$\pm$0.1055}		&0.7881{\scriptsize$\pm$0.0504}	\\
MACMs(poly)   		&61.0028{\scriptsize$\pm$1.3343}	&62.2639{\scriptsize$\pm$1.0108}	&0.8170{\scriptsize$\pm$0.0335}		&0.5981{\scriptsize$\pm$0.0874}	\\
NBMs                       	&56.1521{\scriptsize$\pm$0.8341}	&56.2230{\scriptsize$\pm$0.5366}	&0.8220{\scriptsize$\pm$0.0174}		&0.4732{\scriptsize$\pm$0.0493}	\\
ProtoNAM 		&55.1818{\scriptsize$\pm$0.7710}	&55.2775{\scriptsize$\pm$0.2916}	&\textbf{0.8244{\scriptsize$\pm$0.0281}}		&0.4605{\scriptsize$\pm$0.0220}\\
MACMs(NNs)           	&\textbf{53.4050{\scriptsize$\pm$1.3993}}	&\textbf{52.6411{\scriptsize$\pm$1.8121}}	&0.8211{\scriptsize$\pm$0.0662}		&\textbf{0.4036{\scriptsize$\pm$0.0938}}	\\\hline \hline
ESR          			&49.7763{\scriptsize$\pm$0.8667}	&49.4074{\scriptsize$\pm$0.6692}	&0.7582{\scriptsize$\pm$0.0122}		&0.3417{\scriptsize$\pm$0.0297}	\\
DNN         			&49.0046{\scriptsize$\pm$1.2313}	&49.2913{\scriptsize$\pm$0.5592}	&0.8133{\scriptsize$\pm$0.0209}		&0.3592{\scriptsize$\pm$0.0707}	\\ 	\hline \hline
MP(poly based)		&63.9860{\scriptsize$\pm$0.9295} 	&64.5857{\scriptsize$\pm$1.5560}	&0.8159{\scriptsize$\pm$0.0747}		&0.8834{\scriptsize$\pm$0.0744}	\\
AP(poly based)		&63.9305{\scriptsize$\pm$3.1625}	&65.2596{\scriptsize$\pm$0.7027}	&0.8135{\scriptsize$\pm$0.0324}		&0.7818{\scriptsize$\pm$0.1597}	\\
MP(NNs based)		&56.9609{\scriptsize$\pm$2.1844}	&57.9676{\scriptsize$\pm$2.5309}	&0.8187{\scriptsize$\pm$0.0372}		&0.7153{\scriptsize$\pm$0.0681}	\\
AP(NNs based) 		&60.2258{\scriptsize$\pm$1.9126} 	&59.3957{\scriptsize$\pm$1.2092}	&0.8182{\scriptsize$\pm$0.0436}		&0.5730{\scriptsize$\pm$0.0716}	\\ 	\hline
\end{tabular}
\end{table}

\subsection{Ablation Research}
The purpose of our ablation experiments is to disentangle and analyze the individual contributions of the multiplicative and additive part in MACMs. Specifically, we aim to identify which part is more essential for the overall performance of the model, and to assess the impact of substituting polynomial-based shape functions with neural network parameterizations. The results of the ablation experiments are reported in Table \ref{table:experiment}.

The Multiplicative Part (NNs) and Additive Part (NNs) perform worse than MACMs (NNs) across all datasets, indicating that both parts are indispensable for achieving the high performance of MACMs (NNs), and also demonstrating the success of our strategy in CESR to add an additive part in order to decouple constrained parameters and expand the hypothesis space.

For both MACMs (Poly) and MACMs (NNs), the performance difference between the multiplicative and additive part is not substantial. Specifically, the multiplicative part outperforms on the CA Housing and CA Housing (modified) datasets, while the additive part yields better performance on the Water Quality dataset. This further highlights that the multiplicative part in MACMs should not be regarded solely as an interaction effect; rather, it is jointly constituted by both the multiplicative and additive part.

Across all experiments, MACMs (NNs) achieve better performance than MACMs (Poly), with the improvement being especially significant in regression tasks. This finding suggests that substituting polynomials with more complex fully connected layers effectively boosts the performance of MACMs.

\section{Discussion}
The proposed MACMs provide new insights into the development of interpretable models. While originally motivated by the enhancement of multiplicative models, the framework also reveals an unexpected connection to GAMs. Unlike recent advances in GAMs that primarily focus on refining shape functions within an unchanged hypothesis space, MACMs fundamentally expand the hypothesis space by explicitly incorporating higher-order interactions beyond pairwise terms. This expansion explains the notable performance improvements observed in our experiments and suggests that MACMs can serve as a bridge between multiplicative models and GAM-based approaches.

Despite these advantages, several limitations remain. First, the pursuit of greater model complexity inevitably introduces a trade-off between accuracy and interpretability. The dynamic shape functions in MACMs offer only semi-global interpretability, which, although helpful, falls short of the full transparency provided by classical GAMs. Second, the multiplicative component exhibits sensitivity to the number of features. Without careful adjustment of the scaling factor $k$, the model may become unbalanced, leading to insufficient learning in the multiplicative component while the additive part dominates.

\section{Conclusion and Future Work}
This work introduces a novel interpretable model—Multiplicative-Additive Constrained Models (MACMs), which consist of a multiplicative part and an additive part. By incorporating an additive component into a multiplicative model called Curve Ergodic Set Regression (CESR), we decouple the coefficients of the multiplicative and additive components and thereby expand the model’s hypothesis space. MACMs permit the construction of shape functions using arbitrary mapping forms. In this work, we primarily instantiate MACMs with shape functions parameterized by fully connected neural networks, referred to as MACMs (NNs). Given that MACMs (NNs) include an additive component, we further evaluate their performance in comparison with representative NNs-based GAMs. The experimental results indicate that our proposed model yields a notable improvement over CESR. Furthermore, MACMs (NNs) surpass the state-of-the-art NNs-based GAMs, including NBMs and ProtoNAM, in terms of performance.

In the future, we aim to develop more effective interpretability techniques for MACMs. For example, overlaying raw data samples on dynamic shape function plots could help reveal which patterns between features and outputs are more prominent. Another important direction is to design an automatic adjustment mechanism for the scaling factor. Furthermore, we plan to explore constructing models with two-dimensional shape functions, particularly in the context of MACMs built upon ESR, where the number of parameters could be significantly smaller than that of GAMs. Most importantly, we will enhance MACMs with various optimization strategies, such as grid search for hyperparameter tuning, adding output regularization, or incorporating shape function designs from NBMs and ProtoNAMs, in order to further improve the model’s performance.

%%%%%%%%%%%%%%%%%%%%%%%%%%%%%%%%%%%%%%%%%%%%%%%%%%%%%%%%%%%%
\newpage

\bibliographystyle{plain}
\bibliography{main}

\newpage
\appendix

\section{Appendix}
\subsection{ Additional visualization}

\begin{figure}[h]
  \centering
  \includegraphics[width=\textwidth]{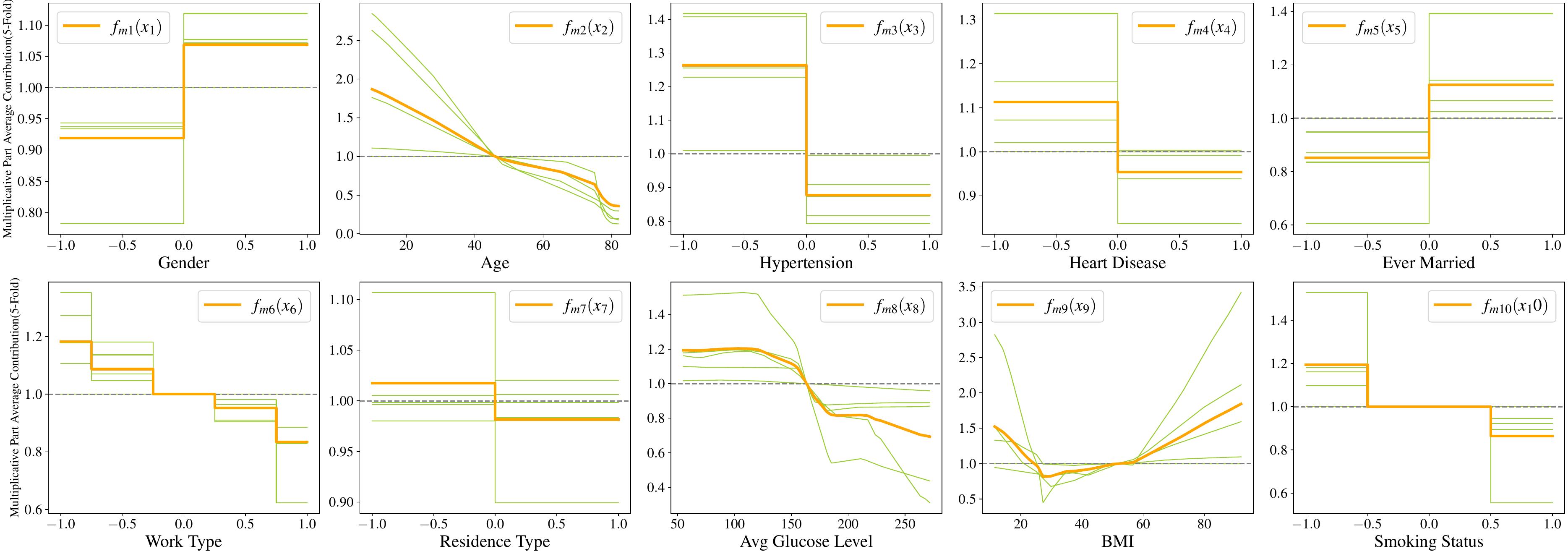}
  \caption{Multiplicative Part Shape Functions of MACMs(NNs) on Stroke Prediction.}\label{figure:mpstroke}
\end{figure}

\begin{figure}[h]
  \centering
  \includegraphics[width=\textwidth]{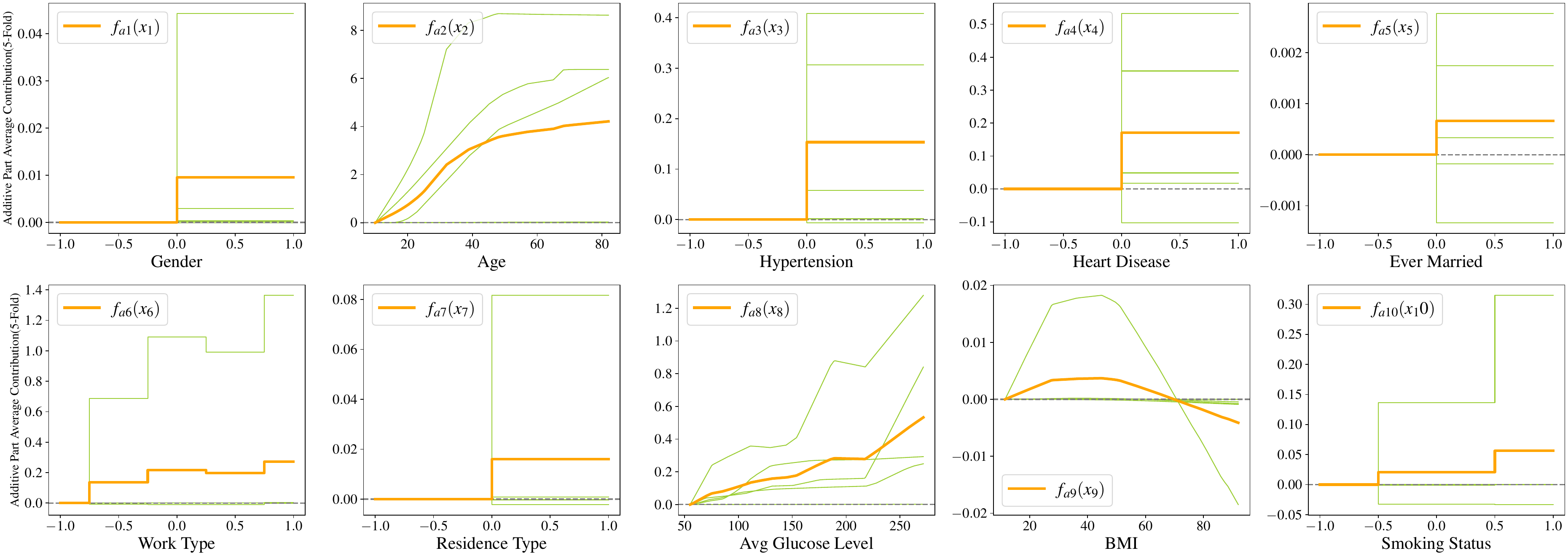}
  \caption{Additive Part Shape Functions of MACMs(NNs) on Stroke Prediction.}\label{figure:apstroke}
\end{figure}

\begin{figure}[h]
  \centering
  \includegraphics[width=\textwidth]{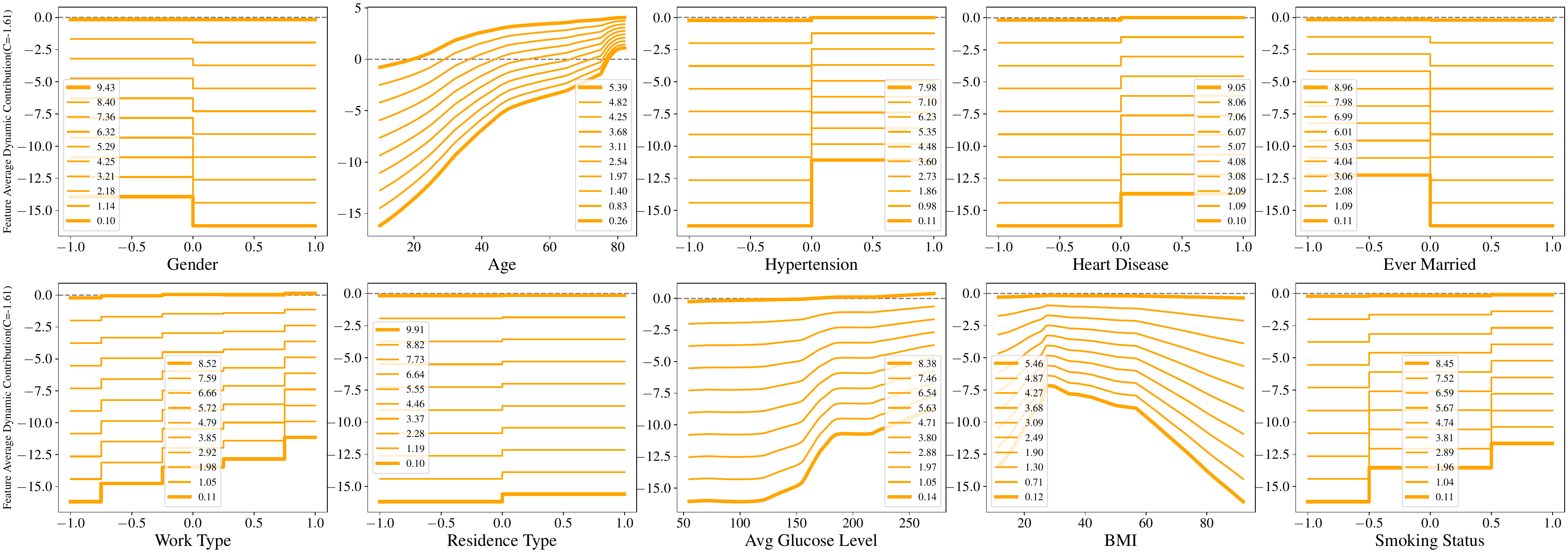}
  \caption{Feature average dynamic contribution shape functions of MACMs(NNs) on Stroke Prediction.}\label{figure:dynamicstroke}
\end{figure}

\begin{figure}[h]
  \centering
  \includegraphics[width=\textwidth]{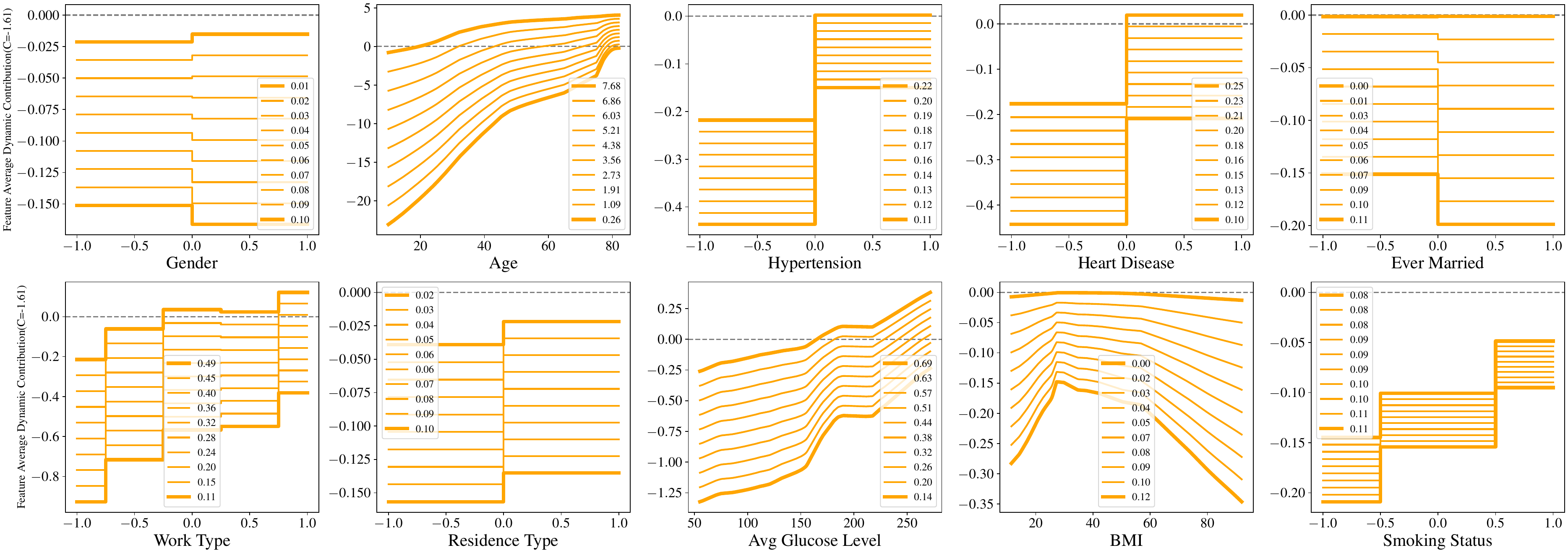}
  \caption{Feature average dynamic contribution shape functions in the first decile interval of MACMs(NNs) on Stroke Prediction.}\label{figure:dynamicstroke10per}
\end{figure}

\end{document}